\documentclass[twocolumn, natbib]{svjour3}

\RequirePackage{fix-cm}
\usepackage{graphicx}%
\usepackage{multirow}%
\usepackage{amsmath,amssymb,amsfonts}%
\usepackage{mathrsfs}%
\usepackage[title]{appendix}%
\usepackage[pagebackref=true,breaklinks=true,citecolor=blue,linkcolor=blue,colorlinks]{hyperref}
\usepackage{xcolor,colortbl}%
\usepackage{textcomp}%
\usepackage{manyfoot}%
\usepackage{booktabs}%
\usepackage{algorithm}%
\usepackage{algorithmicx}%
\usepackage{algpseudocode}%
\usepackage{listings}%

\usepackage{enumitem}   
\usepackage{colortbl}
\usepackage{color}
\definecolor{Gray1}{rgb}{0.92,0.92,0.92}
\definecolor{Gray2}{rgb}{0.88,0.88,0.88}
\definecolor{mediumgray}{rgb}{0.88, 0.88, 0.92}

\usepackage{subcaption}
\newlength\savewidth\newcommand\shline{\noalign{\global\savewidth\arrayrulewidth
  \global\arrayrulewidth 1pt}\hline\noalign{\global\arrayrulewidth\savewidth}}

\makeatletter\renewcommand\paragraph{\@startsection{paragraph}{4}{\z@}
  {.5em \@plus1ex \@minus.2ex}{-.5em}{\normalfont\normalsize\bfseries}}\makeatother

\newcolumntype{x}[1]{>{\centering\arraybackslash}p{#1pt}}
\newcolumntype{y}[1]{>{\raggedright\arraybackslash}p{#1pt}}
\newcolumntype{z}[1]{>{\raggedleft\arraybackslash}p{#1pt}}

\newcommand{\app}{\raise.17ex\hbox{$\scriptstyle\sim$}}

\smartqed

\usepackage{bm}  
\usepackage{xspace} 
\usepackage{booktabs}
\usepackage{adjustbox}
\usepackage{makecell}
\usepackage{mwe}

\newcolumntype{R}[2]{%
    >{\adjustbox{angle=#1,lap=1.3\width-(#2)}\bgroup}%
    l%
    <{\egroup}%
}

\begin{document}\sloppy

\title{PoIFusion: Multi-Modal 3D Object Detection via Fusion at Points of Interest}

\author{Jiajun Deng$^{1}$ \and Sha Zhang$^2$ \and Feras Dayoub$^1$ \and Wanli Ouyang$^3$ \and Yanyong Zhang$^2$ \and Ian Reid$^{1,4}$ }

\authorrunning{Jiajun Deng~\etal} 

\institute{
    Jiajun Deng, corresponding author \at
	\email{jiajun.deng@adelaide.edu.au}           \\
    \\
    Sha Zhang \at
    \email{zhsha1@mail.ustc.edu.cn} \\
    \\
    Feras Dayoub \at
    \email{feras.dayoub@adelaide.edu.au} \\
    \\
    Wanli Ouyang \at
    \email{wanli.ouyang@sydney.edu.au} \\
    \\
    Yanyong Zhang \at
    \email{yanyongz@ustc.edu.cn} \\
    \\
    Ian Reid \at
    \email{ian.reid@mbzuai.ac.ae}\\
    \\
    $^1$The University of Adelaide, Australian Institute for Machine Learning, Adelaide, South Australia, Australia\\
    \\
    $^2$University of Science and Technology of China, Hefei, Anhui, China\\
    \\
    $^3$Shanghai AI Laboratory, Shanghai, China \\
    \\
    $^4$Mohamed bin Zayed University of Artificial Intelligence, Abu Dhabi,  the United Arab Emirates
    }

\date{Received: date / Accepted: date}

\maketitle
\begin{abstract}
In this work, we present PoIFusion, a conceptually simple yet effective multi-modal 3D object detection framework to fuse the information of RGB images and LiDAR point clouds at the points of interest (PoIs). 
Different from the most accurate methods to date that transform multi-sensor data into a unified view or leverage the global attention mechanism to facilitate fusion, our approach maintains the view of each modality and obtains multi-modal features by computation-friendly projection and interpolation.
In particular, our PoIFusion follows the paradigm of query-based object detection, formulating object queries as dynamic 3D boxes and generating a set of PoIs based on each query box.
The PoIs serve as the keypoints to represent a 3D object and play the role of the basic units in multi-modal fusion. 
Specifically, we project PoIs into the view of each modality to sample the corresponding feature and integrate the multi-modal features at each PoI through a dynamic fusion block.
Furthermore, the features of PoIs derived from the same query box are aggregated together to update the query feature.
Our approach prevents information loss caused by view transformation and eliminates the computation-intensive global attention, making the multi-modal 3D object detector more applicable.
We conducted extensive experiments on nuScenes and Argoverse2 datasets to evaluate our approach. Remarkably, the proposed approach achieves state-of-the-art results on both datasets without any bells and whistles, \emph{i.e.}, 74.9\% NDS and 73.4\% mAP on nuScenes, and  31.6\% CDS and 40.6\% mAP on Argoverse2. The code will be made available at \url{https://djiajunustc.github.io/projects/poifusion}.
\keywords{3D Object Detection \and Autonomous Driving \and Multi-Sensor Fusion}
\end{abstract}




\section{Introduction}
\label{sec:intro}

\begin{figure*}[t]
  \centering
  \includegraphics[width=0.98\linewidth]{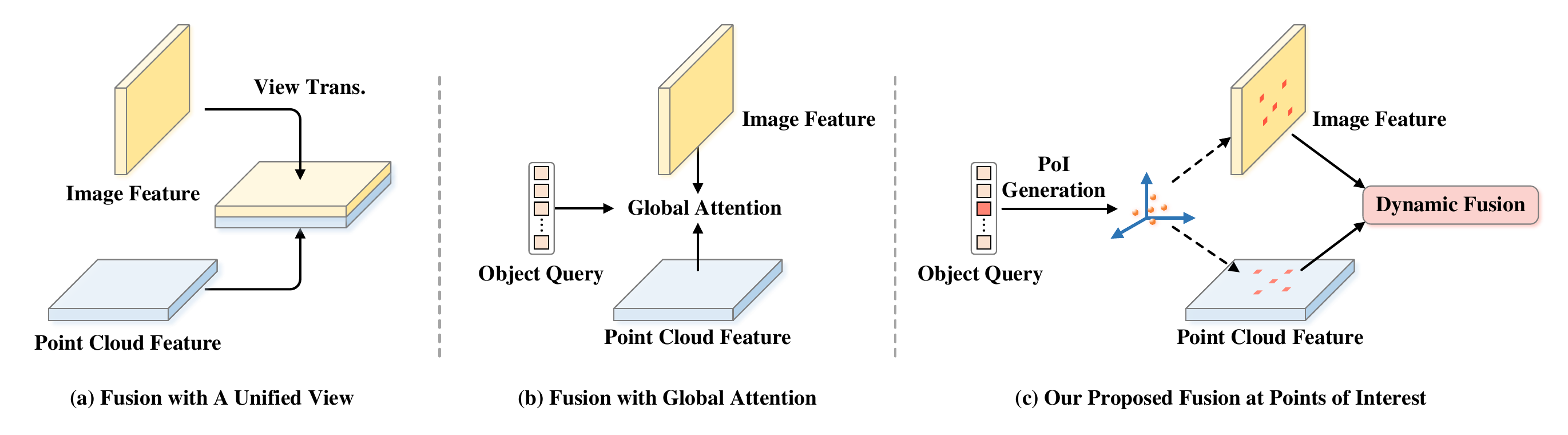}
  \vspace{-0.5cm}
  \caption{A comparison between the representative multi-modal fusion mechanism in the literature and ours: (a) fusion with a unified view, (b) fusion with global attention, and (c) our proposed fusion at points of interest. ``View Trans.'': view transformation. ``PoI'': points of interest.}
  \label{fig:intro}
  \vspace{-0.3cm}
\end{figure*}

Autonomous vehicles are usually equipped with an array of sensors to facilitate safe driving, among which cameras and LiDARs are the most popular. These two sensors are complementary to each other: cameras provide rich textual and color information, while LiDAR sensors supply precise spatial measurements. The effective camera-LiDAR data fusion is widely regarded as a promising direction to achieve high-quality 3D object detection~\cite{transfusion, bevfusion-damo, bevfusion-mit,wang2023multi,zhang2024hvdistill,mao20233d}, which has attracted a surge of research interest in the community.

The fundamental challenge of camera-LiDAR data fusion arises from the discrepancy of their representation space (\emph{i.e.}, 2D perspective view versus 3D space). To ameliorate this challenge, one common solution is to transform the image and point cloud representations into a unified bird-eye view~\cite{bevfusion-mit,bevfusion-damo,metabev}, as depicted in Figure~\ref{fig:intro}(a), or into 3D space~\cite{uvtr}.
Another recent approach~\cite{transfusion,cmt}, as shown in Figure~\ref{fig:intro}(b), keeps the representation in its original view and abstracts multi-modal features into object queries~\cite{detr,petr} with the global attention mechanism~\cite{transformer}.

However, both of these approaches have inherent issues. In the unified-view approach, the core component, \emph{i.e.}, view transformation, is often based on monocular depth estimation to lift the 2D image into 3D. However, depth estimation is an error-prone task, with errors having a deleterious effect on any downstream tasks (such as object recognition).
Moreover, the direct grid-to-grid fusion~\cite{bevfusion-damo,bevfusion-mit} loses a significant portion of the original representational strengths which comprises modal-specific information~\cite{deepinteraction}. 

The second query-based approach can avoid feature ambiguity and information loss by keeping the original view of the feature representation. However, the adoption of global attention to integrating multi-modal features incurs high computation and memory overhead. For instance, the state-of-the-art algorithm~\cite{cmt}, which follows the query-based approach of fusion with global attention, relies on the well-optimized Flash Attention operator~\cite{dao2022flashattention} to cut down the time and memory consumption. The high overhead has become an obstacle that hinders the wide application of the algorithm. Furthermore, it is difficult to extract the object-relevant feature with the global attention mechanism~\cite{deformable_detr,gao2022adamixer}, especially when it comes to such a large 3D space like the autonomous driving scenario. 

The benefit of the query-based approach inspires us to preserve the original view of features from each modality, while the above two issues motivate our exploration of an alternative paradigm to replace dense feature interaction of the global attention with sparse point projection and feature sampling, as illustrated in Figure~\ref{fig:intro}(c).

As such, we propose a new query-based multi-modal 3D object detection framework that initializes object queries as learnable 3D boxes and dynamically integrates multi-modal features at representative points derived from each object query. In this manuscript, these representative points are referred to as \textbf{Points of Interest} (PoIs), and our framework is named \textbf{PoIFusion}. Intuitively, a naive way to represent a 3D query box with points is to use the center point~\cite{futr3d,cmt,petr}. However, simply representing a 3D box with its center point totally ignores the geometric properties, such as the size and rotation angle. For supplementary, an improved design is to involve the corners~\cite{sheng2021improving}. Nevertheless, sampling multi-modal features according to the projected location of center and corner points also incurs the problem of feature misalignment, since the projection of a 3D box may not be a tight bounding box onto the image view. The issue remains even if a box is well-located in the 3D space~\cite{cai2023objectfusion}, not to mention that the query box is not guaranteed to be accurately localized.
To ameliorate this problem, our PoIs are adaptively generated from the center and corner points, with box-level and point-level transformation parameters online predicted according to the query feature.
In our design, a single PoI serves as the basic unit for fine-grained multi-modal feature fusion, and the ensemble of PoIs derived from the same object query represents the regional feature in a flexible way.

Once PoIs are generated, the features from different modalities can be easily obtained by projecting PoIs onto the corresponding view, followed by bilinear interpolation according to the projected location. Moreover, because different modalities contribute differently to each object query, a dynamic fusion block is engaged in our design. Particularly, in our dynamic fusion block, we first generate the parameters of the fusion layer on the fly, and then integrate the sampled multi-modal feature at each PoI. The proposed adaptive PoIs, together with the dynamic fusion block, enable our PoIFusion to efficiently sample the object-relevant multi-modal features and make the best of modal-specific information, thus improving 3D object detection.

We evaluate the proposed PoIFusion framework on the nuScenes and Argoverse2 datasets, conducting a comprehensive set of experimental analyses to validate our design choices. Notably, PoIFusion achieves state-of-the-art performance on the highly competitive nuScenes benchmark, attaining 74.9\% NDS and 73.4\% mAP on the test set without any bells and whistles. Moreover, when applied to the more challenging Argoverse2 dataset, our method achieves 40.6\% mAP and 31.6\% CDS, improving the best performance in the literature by absolutely 4.5\% mAP and 3.8\% CDS.

In summary, we make three-fold our contributions:
\begin{itemize}[leftmargin=2em]
  \item We propose a novel PoIFusion framework for multi-modal 3D object detection that preserves modality-specific representation spaces while efficiently extracting and fusing features through sparse interactions.
  \item We present the design of fusion at points of interest, conveying an elegant view that the entity involved in the fusion module can be very flexible.
  \item We conduct extensive experiments to validate the effectiveness of our method, demonstrating its potential to serve as a strong baseline for this field.
\end{itemize}

\section{Related Work}
\label{sec:related_work}

\subsection{LiDAR-Based 3D Object Detection.}
LiDAR sensors capture 3D point clouds, which provide accurate spatial information that can be important for 3D object detection. Broadly, 3D object detection algorithms operated on point clouds can be categorized into two groups: point-based and voxel-based ones. Point-based 3D object detection algorithms~\cite{pointrcnn,3dssd,std} make direct use of the precise coordinates of points, progressively sampling keypoints and extracting local information with set abstraction operators~\cite{pointnet,pointnet++,pvrcnn,pvrcnnpp} that aggregate information form a point and its neighbors. In contrast, voxel-based 3D object detection algorithms~\cite{voxelnet,yang2022towards,pointpillar,voxelrcnn,centerpoint,pillarnet,fsd,deng2021multi,yang2022st3d++,wang2023bilrfusion,wang2024club} first ``voxelize'' the point cloud by binning points into a regular grid. The voxel representation enables straightforward feature extraction using standard (or sparse) convolutions~\cite{second,focalspconv,largekernel,parta2} or sparse voxel Transformers~\cite{sst,wang2023dsvt,centerformer,lai2023spherical,chen2023focalformer3d,dong2022mssvt,mao2021voxel}.

\begin{figure*}[t]
  \centering
  \includegraphics[width=0.99\linewidth]{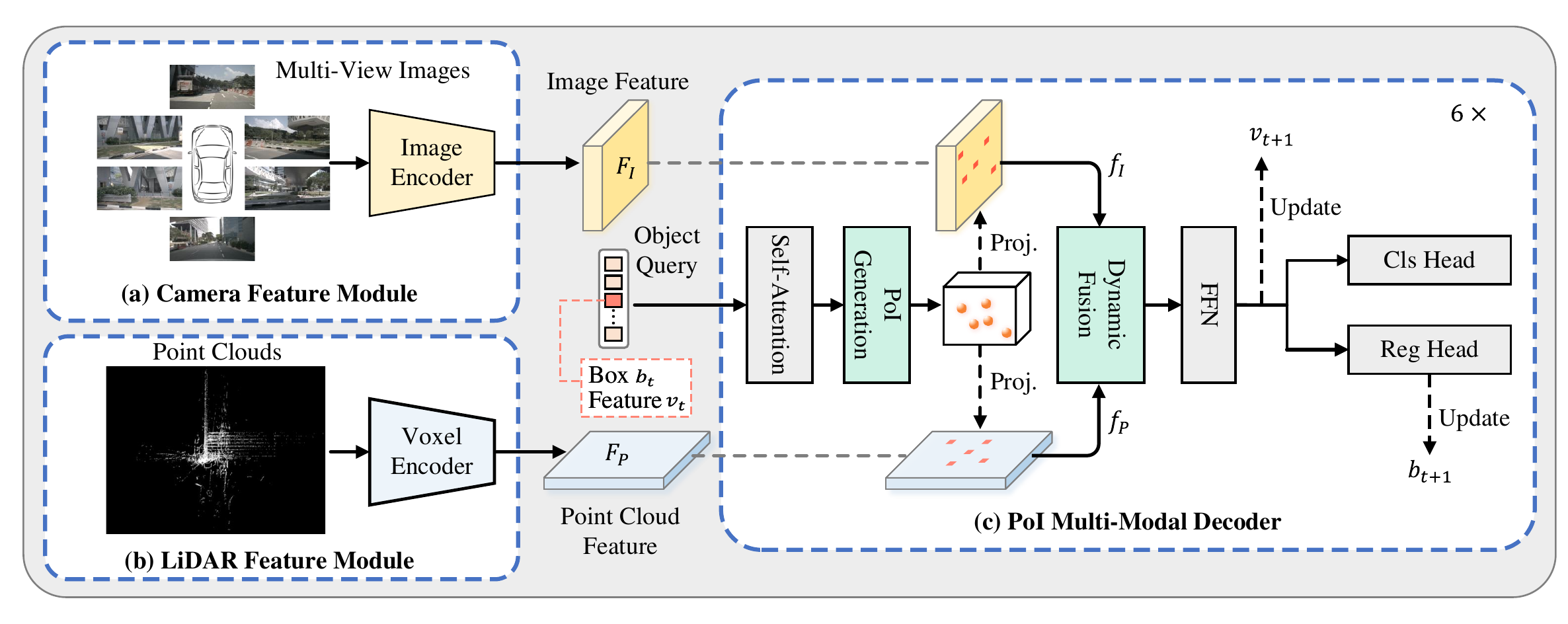}
  \caption{An overview of our proposed PoIFusion framework, which is mainly composed of (a) a camera feature module, (b) a LiDAR feature module, and (c) a PoI multi-modal decoder. In our method, we first independently extract the feature of each modality and keep the original representation view (\emph{i.e.}, image feature in the perspective view, and point cloud feature in the bird-eye view). The multi-modal feature maps are taken as the input of our PoI multi-modal decoder. The decoder is iteratively applied 6 times to integrate the multi-modal feature sampled with generated points of interest (PoIs) and to refine the object queries. In this figure, ``proj.'' stands for ``projection''.
  } 
  \label{fig:framework}
\end{figure*}

\subsection{Camera-Based 3D Object Detection.}
3D object detection is one of the oldest Computer Vision problems and a variety of approaches have been proposed over many years (we do not survey these here). For the purposes of 3D object detection from a moving car a popular approach, and one that can be readily fused with point-cloud estimates, is to measure detection success in the Birds-eye View (BEV) space. Earlier attempts along these lines -- \emph{e.g.}, geometric uncertainty~\cite{Lu_2021_ICCV} and pseudo-LiDAR representation~\cite{wang2019pseudo} -- mainly focus on monocular 3D object detection~\cite{fcos3d,caddn,brazil2019m3d,chong2022monodistill,zhang2023monodetr,zhou2021monoef}. Autonomous driving vehicles are typically equipped with multiple cameras~\cite{xie2021vote, guo2024cyclic}, providing perception information over the full 360 degrees. To leverage the relationship between multi-view images, BEV-based and query-based algorithms have been explored. BEV-based 3D object detection~\cite{huang2021bevdet,bevformer,bevdepth,li2023delving} explicitly performs view transformation to unify multi-view images into bird-eye-view representation. Query-based 3D object detection~\cite{petr,detr3d,liu2023sparsebev,wang2023object,shu20233dppe,xiong2023cape} follows the pipeline of DETR~\cite{detr}, capitalizing on object queries to extract multi-view information without view transformation.

\subsection{Multi-Modal 3D Object Detection.}
Although the exploration of each individual modality has made encouraging progress, the accuracy and robustness of detection algorithms are still insufficient for safe driving. The fact that images and point clouds are naturally complementary to each other (\emph{i.e.}, rich semantic information versus precise spatial information) has motivated further exploration in multi-modal 3D object detection~\cite{wang2023unitr,transfusion,futr3d,metabev,sparsefusion,li2023fully,jiao2023msmdfusion,yin2024fusion}. In the early stage, the multi-modal approaches utilize point clouds as the principal component, introducing image features at the point level~\cite{pointaugmenting,pointpainting,mvp,autoalign} or proposal level~\cite{zhu2022vpfnet,mv3d,3d-cvf} to enhance the features of the point clouds. Recently, inspired by multi-view 3D object detection, a series of works~\cite{bevfusion-damo,bevfusion-mit,uvtr} propose that unifying the representation space with explicit view transformation~\cite{lss} facilitates multi-modal fusion.
Another group of methods~\cite{deepfusion,transfusion,cmt,deepinteraction} leverages the attention mechanism in Transformer architecture~\cite{transformer} to perform multi-modal fusion in a sequential~\cite{transfusion} or a parallel~\cite{cmt} manner. 
A recent work, ObjectFusion~\cite{cai2023objectfusion}, fuses multi-modal features in a two-stage pipeline.
It first generates region proposals with the image-augmented BEV features~\cite{transfusion}, and then extracts the object-centric feature~\cite{maskrcnn,voxelrcnn} from the voxel, image, and BEV space for further fusion and refinement. 

In this work, the proposed PoIFusion scheme is also object-centric. 
However, in contrast to integrating the corresponding region-wise feature from multiple modalities, we adaptively generate PoIs from each object query, and leverage the PoIs as the basic units to perform multi-modal fusion - this element stands as the cornerstone of our approach, fundamentally improving the efficacy and flexibility of our multi-modal 3D object detection framework.

\section{Our Approach}

In this section, we present the details of our PoIFusion.
In Section~\ref{sec3:overview}, we provide a comprehensive overview of our framework. Then, in Section~\ref{sec3:poi}, we explain how to generate PoIs based on the object query. 
Subsequently, in Section~\ref{sec3:sampling}, we introduce the process of multi-modal feature sampling. After that, we elaborate on our dynamic fusion block design in Section~\ref{sec3:fusion}. Finally, in Section~\ref{sec3:pred}, we detail our prediction head and the training objective.

\subsection{Overview} \label{sec3:overview}
As illustrated in Figure~\ref{fig:framework}, our PoIFusion consists of three main components: (a) a camera feature module, (b) a LiDAR feature module, and (c) a PoI multi-modal decoder. 
Given multi-view images and point clouds, perspective-view image feature maps $\bm{F}_I$ and bird-eye-view (BEV) point cloud feature maps $\bm{F}_P$ are independently extracted with the image encoder and the voxel encoder. The image encoder is composed of an image backbone network and an FPN~\cite{fpn} to obtain multi-scale features. The voxel encoder exploits a sparse 3D backbone network~\cite{second,voxelnet} and a BEV backbone network, following the common practice of the voxel-based paradigm.
Then, the PoI multi-modal decoder, which works as the core component in our method, is iteratively applied 6 times, progressively integrating the multi-modal features and refining the detection boxes.

In our multi-modal decoder, each object query $\bm{Q}$ is formulated as an adaptive 3D box $\bm{b}_t$, associating with a feature vector $\bm{v}_t$ ($t$ indicates the iteration time step). 
In each iteration, the object queries are first fed into a self-attention layer to capture the relationships and dependencies between different objects. Here, we follow~\cite{liu2023sparsebev} to exploit a distance-biased self-attention layer in our decoder.
Subsequently, for each object query, the box-wise transformation parameters and point-wise shift parameters are generated based on the query feature $\bm{v}_t$, and applied on the query box $\bm{b}_t$ to obtain a set of PoIs $\bm{P}_t=\{P^i\}$.
To perform multi-modal fusion, a three-step process is undertaken for PoIs. Firstly, each PoI is projected onto both the perspective view and the bird-eye view, establishing the correspondence between the PoI and the multi-modal feature maps. Secondly, the image feature $\bm{f}_I$ and point cloud feature $\bm{f}_P$ of a PoI are sampled through bilinear interpolation, regarding the projected location on the corresponding view. Thirdly, the extracted multi-modal features at each PoI are dynamically integrated using a dynamic fusion block, followed by a feedforward network (FFN). Finally, a classification head and a regression head are applied for prediction, and the query feature and query box are updated, respectively.

\subsection{Object Query Initialization.} 
We formulate each object query as a learnable dynamic 3D box $\bm{b}\in\mathbb{R}^8$ and an attached feature vector $\bm{v}\in\mathbb{R}^{256}$. Specifically, the query box is represented as:
\begin{flalign}
&&
\bm{b} =  [x_c, y_c, z_c, w, l, h, sin\theta,cos\theta],
&&
\end{flalign}
where [$x_c$, $y_c$, $z_c$] is the center location, [$w$, $l$, $h$] is the box dimension, and $\theta$ is the azimuth angle.  Notably, although velocity is also predicted in the detection results, it is not factored into the query boxes. 

Prior to training, the object query is initialized as follows: the BEV center location [$x_c$, $y_c$] is set to be uniformly distributed on the BEV space, and the height center $z_c$ is set as 0. Besides, the dimension triplet [$w$, $l$, $h$] of each box is initialized as [6, 3, 2], which is about the average size of objects calculated on the dataset.  and the azimuth is initialized as 0. The corresponding attached feature vector is randomly initialized.

\subsection{PoI Generation} \label{sec3:poi}

In this section, we elaborate on how to generate points of interest (PoIs), which serve as the basic units for multi-modal feature fusion in our PoIFusion framework.

Let us consider one object query $\bm{Q}=\{\bm{b},\bm{v}\}$ as an example, where $\bm{b}$ is the 3D query box and $\bm{v}$ is the query feature (the iteration step t is omitted in the following chapters).
The spatial information of $\bm{b}$ is presented as [$x_c$, $y_c$, $z_c$, $w$, $l$, $h$, sin$\theta$, cos$\theta$], where [$x_c$, $y_c$, $z_c$] is the center location, [$w$, $l$, $h$] is the box dimension, and $\theta$ is the heading direction in bird-eye view.
In the PoI generation block, two sibling linear layers are applied to $\bm{v}$, producing box-wise transformation parameters $\Delta_B=\text{[}t_x, t_y, t_z, t_w, t_l, t_h, t_\text{sin}, t_\text{cos}\text{]}$ and point-wise shift parameters $\{\Delta_P^i=\text{[}\delta_x^i, \delta_y^i, \delta_z^i\text{]}\}_{i=0}^8$. Subsequently, a holistic box transformation according to $\Delta_B$ is performed on query box $\bm{b}$ to obtain transformed box $\bm{b}'$:
\begin{flalign}
&&
x_c' &= x_c + t_x, & w' &= w \cdot \exp(t_w), \\
&&
y_c' &= y_c + t_y, & l' &= l \cdot \exp(t_l), \\
&&
z_c' &= z_c + t_z, & h' &= h \cdot \exp(t_h), \\
&&
\sin\theta' &= \sin\theta + t_{\sin}, & \cos\theta' &= \cos\theta + t_{\cos}.
&&
\end{flalign}
Once the box transformation has been applied, the center point and 8 corner points of the transformed box $\bm{b'}$ are collected together as the anchor points $\{A^i\}_{i=0}^8$. Finally, the point-wise shift is independently applied to each anchor point $A^i$ to produce a PoI $P^i$ ($P^i=A^i+\Delta_P^i$). 
It's worth noting that one object query corresponds to a set of PoIs derived from the center and corner points of the query box.

\subsection{Feature Sampling} \label{sec3:sampling}

In our approach, we assemble multi-modal features by establishing the correspondence between a PoI and multi-modal feature maps via projection, followed by sampling features utilizing bilinear interpolation.

Let us denote the location of a 3D PoI $P^i$ as [$x^i$, $y^i$, $z^i$]. To sample the point cloud feature, $P^i$ is projected onto the bird-eye view (BEV). Since we exploit the voxel representation for point clouds, the location of the projected point on the BEV point cloud feature $F_P$ is computed as:
\begin{flalign}
    && 
    \begin{aligned}
        \begin{cases}
            m^i & = \dfrac{x^i - X_{\text{min}}}{\left( X_{\text{max}} - X_{\text{min}} \right) \times d}, \\[2ex]
            n^j & = \dfrac{y^i - Y_{\text{min}}}{\left( Y_{\text{max}} - Y_{\text{min}} \right) \times d},
        \end{cases}
    \end{aligned}
    &&
\end{flalign}
where [$m^i$, $n^i$] is the projected location, [$X_\text{min}$, $Y_\text{min}$, $X_\text{max}$, $Y_\text{max}$] is the point cloud range in the BEV, and $d=8$ is the downsampling scalar of point cloud feature maps. The point cloud feature sampled at [$m^i$, $n^i$] is denoted as $\bm{f}^i_P$.

Typically, there are multiple surrounding-view cameras on the autonomous vehicle, resulting in multi-view images. A 3D point can be projected onto one or two views. If the PoI is projected to only one view, this view will be leveraged to perform image feature sampling. Otherwise, we follow~\cite{cai2023objectfusion} to randomly choose one view for image feature sampling.
Given a 3D PoI $P^i$ in the LiDAR coordinate system, the projection onto the 2D image plane can be computed as follows:
\begin{flalign}
&& 
P_{\text{image}}^i = \mathcal{I} \cdot (\mathcal{R} \cdot P^i + \mathcal{T}),
&&
\end{flalign}
where $P_{\text{image}}^i$ represents the coordinates of the projected PoI on the image plane, $\mathcal{I}$ is the intrinsic camera matrix, $\mathcal{R}$ is the rotation matrix of the extrinsic parameters, and $\mathcal{T}$ is the translation vector of the extrinsic parameters.

In our image encoder, we exploit an image backbone network followed by an FPN~\cite{fpn} to produce feature maps at P2, P3, P4, and P5 (1/4, 1/8, 1/16, and 1/32 downsampling, respectively). 
For a given projected point $P_{\text{image}}^i$, a set of image feature $\{\bm{g}^i_j\}_{j=2}^5$ is sampled from the multi-scale feature maps via bilinear interpolation.
The sampled multi-scale image features are then aggregated as:
\begin{flalign}
    && 
\bm{f}_I^i =  \frac{\sum_{j=2}^5 \text{exp}(w_j^i) \cdot \bm{g}_j^i}{\sum_{j=2}^5 \text{exp}(w_j^i)},
&&
\end{flalign}
where $\bm{f}_I^i$ represents the aggregated image feature and $\{w^i_j\}_{j=2}^5$ are scale weight coefficients, which are predicted by a linear layer with the query feature $\bm{v}$.

\subsection{Dynamic Multi-Modal Feature Fusion} \label{sec3:fusion}
After feature sampling, each PoI is attached with a multi-modal feature pair. The next step is to fuse the multi-modal feature and integrate it into the object query. One simple solution is to exploit a static linear layer over the concatenated feature pair for fusion. However, using static linear layers ignores the fact that the image and point cloud features contribute differently to different objects. To this end, our fusion block capitalizes on a dynamic fusion scheme.

\begin{figure}[t]
  \centering
  \includegraphics[width=0.85\linewidth]{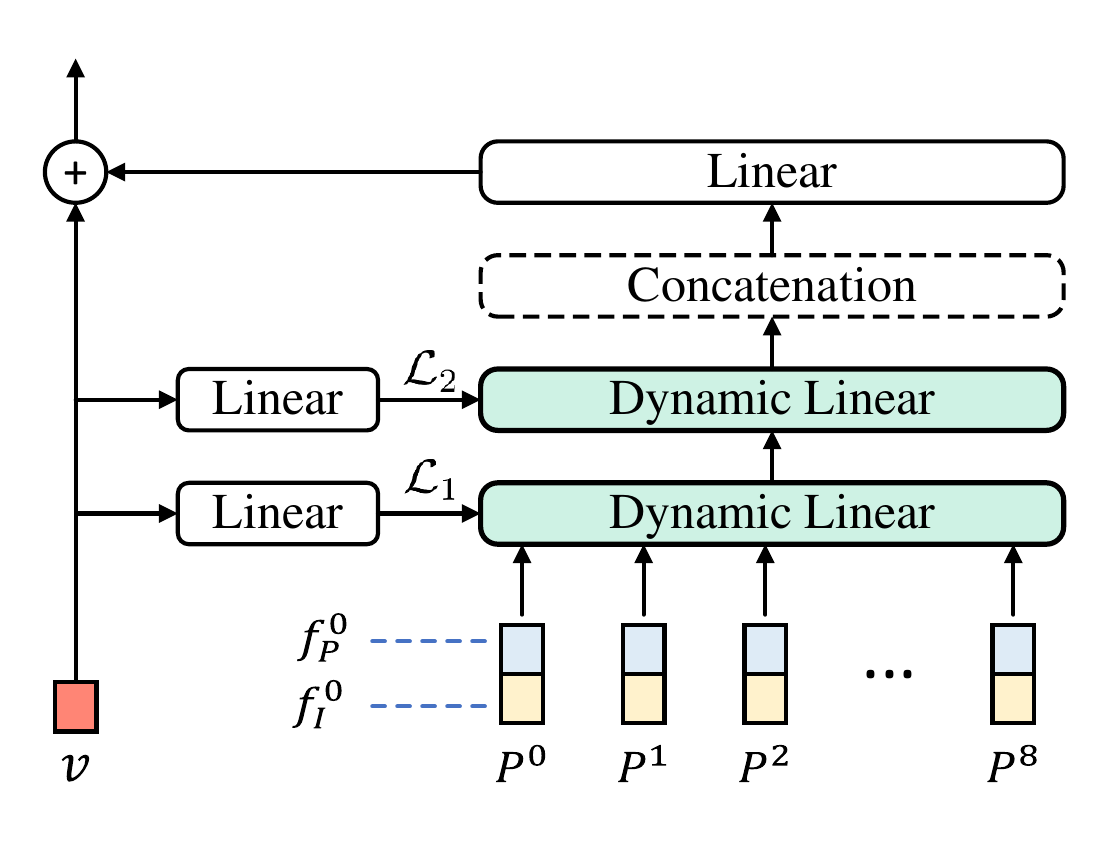}
  \caption{An illustration of our dynamic multi-modal feature fusion block, which first fuses image and point cloud feature at each PoI, and then integrates the PoIs of the same object query in canonical order.}
  \label{fig:fusion}
\end{figure}

The detail of our dynamic fusion block is depicted in Figure~\ref{fig:fusion}. After PoI generation and feature sampling, each object query $\bm{Q}$ is represented by a set of PoIs $\{P^i\}$, and each PoI is attached with a sampled multi-modal feature pair $\{\bm{f}_P^i, \bm{f}_I^i\}$.
Firstly, we exploit conventional linear layers to produce dynamic fusion parameters $\mathcal{L}_1$ and $\mathcal{L}_2$ according to the query feature $\bm{v}$. After that, the concatenated multi-modal feature pair of each PoI is individually integrated through the dynamic linear layers parameterized by $\mathcal{L}_1$ and $\mathcal{L}_2$. Note that the distinction between the dynamic linear layer and the conventional linear layer is that the parameters of the dynamic one are produced on the fly. Subsequently, the features of PoIs derived from the same object query are concatenated together and fed into an additional linear layer to perform PoI feature aggregation. Finally,  the aggregated feature is added back to the query feature. The dynamic linear layers and the conventional linear layer leveraged for PoI feature aggregation are followed by a Layer Normalization operation~\cite{ba2016layer} and a ReLU activation layer.

\subsection{Prediction Head and Training Objective} \label{sec3:pred}
 
\noindent\textbf{Prediction Head.} 
Our prediction head consists of a classification head and a sibling box regression head. In the classification head, we predict the binary classification score for each category. In the regression head, we follow the iterative refinement scheme~\cite{teed2020raft,deformable_detr,lin2023detr} to predict the center delta over the predicted box center from the previous iteration as follows:
\begin{flalign}
    && 
\bm{c}_t = \bm{c}_{t-1} + \Delta \bm{c}_t,
&&
\end{flalign}
 where $\bm{c}_t$ is the predicted box center predicted in the $t$-th recurrent iteration  and $\Delta \bm{c}_t$ is the center delta.
 The other parts of the predicted 3D bounding box, including the box dimension, the heading direction, and the velocity, are independently predicted at each time.

\vspace{+0.1cm}
\noindent\textbf{Training Objective.}
We follow the practice of set prediction~\cite{detr} for the target assignment. Specifically, bipartite matching is exploited to obtain one-to-one assignments between predicted results and ground-truth ones. In corresponding to the prediction head, our training objective also includes two parts: a focal loss~\cite{focalloss} for the classification head and an L1 loss for the regression head. The overall training objective is computed as follows:
\begin{flalign}
    && 
L = \alpha \cdot L_{cls} (\bm{y}_{gt}, \bm{y}_{pred}) + \beta \cdot L_{reg} (\bm{b}_{gt}, \bm{b}_{pred})
    && 
\end{flalign}
where $\bm{y}$ indicates classification logits, $\bm{b}$ means box coordinates,
and $\alpha$ and $\beta$ are coefficients that balance these two losses. We set $\alpha$ to 2.0 and set $\beta$ to 0.25 following~\cite{cmt,petr}.

\section{Experiments}

\begin{table*}[t]
\centering
\small
\renewcommand{\arraystretch}{1.2}
\setlength\tabcolsep{5.2pt}
\caption{
\textbf{Performance comparison with state-of-the-art methods on nuScenes validation/test split.} 
C: camera data (RGB images). 
L: LiDAR data (point clouds). 
$\star$: the image input depends on predicted 2D instance masks.
$\triangle$: the image resolution of CMT is enlarged to $1600\times 640$.
Our approach does not utilize test-time augmentation or employ model ensembling. 
}
\resizebox{0.99\textwidth}{!}{%
\begin{tabular}{l|c|c|c|cc|cc}
\specialrule{1pt}{0pt}{1pt}
\toprule
\multirow{2}{*}{\textbf{Methods}} & \multirow{2}{*}{Modality} & \multirow{2}{*}{Camera Backbone} & \multirow{2}{*}{LiDAR Backbone} & \multicolumn{2}{c|}{\textit{val}} & \multicolumn{2}{c}{\textit{test}} \\
                        &                            &            &     & mAP~$\uparrow$ & NDS~$\uparrow$   & mAP~$\uparrow$ & NDS~$\uparrow$ \\ \midrule
DETR3D~\cite{detr3d}                  & C                         & ResNet-50 & - & 34.9       & 43.4        & 41.2       & 47.9 \\
BEVFormer~\cite{bevformer}               & C                         & ResNet-50 & - & 41.6       & 51.7        & 48.1       & 56.9 \\ \midrule
CenterPoint~\cite{centerpoint}             & L                         & - & VoxelNet & 59.6       & 66.8        & 60.3       & 67.3 \\
TransFusion-L~\cite{transfusion}           & L                         & - & VoxelNet & 65.1       & 70.1        & 65.5       & 70.2 \\
\midrule
MVP$^\star$ ~\cite{mvp}                     & C+L                       & ResNet-50 & VoxelNet & 67.1       & 70.8        & 66.4       & 70.5 \\
PointAugmenting~\cite{pointaugmenting}         & C+L                       & ResNet-50 & VoxelNet & -          & -           & 66.8       & 71.0 \\
FUTR3D~\cite{futr3d}   & C+L&ResNet-101 & VoxelNet                                   & 64.5       & 68.3        & -          & -          \\
TransFusion~\cite{transfusion}             & C+L                       & ResNet-50 & VoxelNet & 67.5       & 71.3        & 68.9       & 71.6 \\
UVTR~\cite{uvtr}             & C+L                       & ResNet-101 & VoxelNet & 65.4       & 70.2        & 67.1       & 71.1 \\
BEVFusion (PKU)~\cite{bevfusion-damo}               & C+L                       & Dual-Swin-T & VoxelNet & 69.6       & 72.1        & 71.3       & 73.3 \\
DeepInteraction~\cite{deepinteraction}         & C+L                       & ResNet-50 & VoxelNet & 69.9       & 72.6        & 70.8       & 73.4 \\
BEVFusion (MIT)~\cite{bevfusion-mit}               & C+L                       & Swin-T & VoxelNet & 68.5       & 71.4        & 70.2       & 72.9 \\
MSMDFusion$^\star$~\cite{jiao2023msmdfusion} & C+L & ResNet-50 & VoxelNet & 69.3 & 72.0 & 71.5 & 74.0 \\
UniTR~\cite{wang2023unitr} & C+L & Multi-Modal DSVT &  Multi-Modal DSVT& 70.0 & 73.1 & 70.5 & 74.1 \\
SparseFusion~\cite{sparsefusion}           & C+L                       & ResNet-50 & VoxelNet & 70.4       & 72.8      & 72.0       & 73.8 \\
CMT$^\triangle$~\cite{cmt}              & C+L        & VoVNet-99 & VoxelNet & 70.3       & 72.9      & 72.0       & 74.1 \\ 
ObjectFusion~\cite{cai2023objectfusion}              & C+L              & Swin-T & VoxelNet            & 69.8       & 72.3      & 71.0       & 73.3 \\
FSF$^\star$~\cite{li2023fully}            & C+L                       & ResNet-50 & Sparse-ResUNet & 70.4       & 72.7      & 70.6       & 74.0 \\
\midrule
\rowcolor{mediumgray} PoIFusion (ours)              & C+L                       & ResNet-50 & VoxelNet & 71.2      & 73.2      & 72.6     & 74.3 \\
\rowcolor{mediumgray} PoIFusion (ours)              & C+L                       & Swin-T & VoxelNet & \textbf{71.7}       & \textbf{73.6}      & \textbf{73.4}      & \textbf{74.9} \\
\bottomrule
\specialrule{1pt}{1pt}{0pt}
\end{tabular}%
}
\label{tab:overall}
\end{table*}

\subsection{Dataset and Evaluation Metric}
We evaluate our approach on nuScenes~\cite{nus} and Argoverse2~\cite{argo2} datasets. 

\noindent \textbf{NuScenes.} The nuScenes dataset includes 700, 150, and 150 driving scenes for training, validation, and testing, respectively. 
The vehicle for data collection is equipped with a 32-beam LiDAR sensor and 6 surrounding-view RGB cameras, thus providing both point clouds and multi-view images.
Besides, this dataset annotates more than 1.4 million 3D bounding boxes, across 10 common categories on the street. Both the multi-sensor data and annotation facilitate the exploration of multi-modal 3D object detection. 
We follow the official policy to mainly evaluate our method on the 3D object detection benchmark in terms of mean average precision (mAP) and nuScenes detection score (NDS). 
The mAP is averaged over distance thresholds $0.5m$, $1m$, $2m$, and $4m$ on the BEV across $10$ classes. NDS is a weighted average of mAP and other true-positive metrics including mATE, mASE, mAOE, mAVE, and mAAE.

\noindent \textbf{Argoverse2.} The Argoverse 2 (AV2) dataset is a large-scale benchmark for perception and prediction in autonomous driving. It comprises 150,000 annotated frames, which is five times larger than the nuScenes dataset, and 1,000 driving scenes. It features two 32-beam LiDARs combined into a 64-beam LiDAR and seven high-resolution surrounding cameras, offering a full 360° field of view and a valid detection range of up to 200 meters, covering an area of 400m × 400m. The dataset is particularly suited for long-range multi-modal object detection tasks. We evaluate our method across 26 categories. For evaluation, in addition to mean Average Precision (mAP), AV2 provides the Composite Detection Score (CDS), a comprehensive metric that combines mAP with other true positive metrics like Average Translation Error (mATE), Average Scale Error (mASE), and Average Orientation Error (mAOE).

\subsection{Experimental Setup}\label{sec4.1:exp_setup}
\noindent\textbf{Network Configuration.}
The image encoder is composed of an image backbone network (\emph{i.e.}, ResNet~\cite{resnet} or Swin-Transformer~\cite{swin}) and an FPN~\cite{fpn} to enrich the multi-scale information. 
The voxel encoder comprises of a 3D voxel backbone network and a BEV backbone network.
For nuScenes, we follow the common practice of using VoxelNet~\cite{voxelnet} as the 3D backbone network, setting the image resolution as $800\times448$, and setting the voxel size as ($0.075m$, $0.075m$, $0.2m$). The number of object queries is set as $900$. For Argoverse2, there is less literature that can be referred to. Therefore, we follow FSF~\cite{li2023fully} to use SparseResUNet as the 3D backbone network. The image resolution is $960\times640$ and the voxel size is ($0.2m$, $0.2m$, $0.2m$). It is worth noting that FSF keeps the output of SparseResUNet without downsampling, while we downsample the output for 4 times to decrease the resolution of BEV features. Since applied for a larger detection range, we use $1600$ object queries on Argoverse2.
The configuration of our PoI multi-modal detector on both datasets keeps the same.
Specifically, the feature dimension of query embedding is set as $256$. The channel of image and point cloud features is also transformed to $256$ before being fed into the multi-modal decoder. We equally divide the channels of feature maps of each modality into 4 groups, and generate PoIs for each group. This grouping operation improves the capacity of the network~\cite{xie2017aggregated} and reduces the computation costs of the dynamic fusion block. Moreover, the first dynamic fusion layer halves the channel of the query embedding, and the second dynamic fusion layer restores the feature dimension. The PoI multi-modal decoder is iteratively applied 6 times, with shared parameters.

\vspace{+0.1cm}
\noindent\textbf{Training and Inference.}
To make fair comparison, we follow the same training pipeline of previous methods~\cite{bevfusion-mit,cai2023objectfusion,sparsefusion,bevfusion-damo,deepinteraction,li2023fully}, and use the same pretrained models. Specifically, the image encoder is pre-trained on nuImage~\cite{nus} dataset. For nuScenes, the voxel encoder is initialized with the model weights of pretrained TransFusion-L~\cite{transfusion}. For Argoverse2, the 3D voxel backbone network is initialized with the model weights of pretrained FSD~\cite{fsd}, while the BEV backbone network is randomly initialized. For both datasets, our fusion framework is trained for 6 epochs with the AdamW optimizer. For nuScenes, the training sample is resampled by CBGS~\cite{cbgs} strategy. The initial learning rate is set as 1e-4, adapted with the one-cycle learning rate policy, and the weight decay is set as 0.01. 
Data augmentation including random flip, random rotation, random translation, random scaling, and random modal masking is adopted. We exploit 8 GPUs for training, with 2 samples on each GPU for nuScenes and 1 sample on each GPU for Argoverse2.

At inference, PoIFusion outputs the top 300 detection boxes without NMS. We don't exploit any test time augmentation or model ensemble techniques.

\begin{table*}[!ht]
\setlength{\tabcolsep}{3pt}
\renewcommand{\arraystretch}{1.2}
\caption{
\textbf{Performance comparison with state-of-the-art methods on Argoverse 2 validation split.}
C-Barrel: construction barrel.
MPC-Sign: mobile pedestrian crossing sign.
A-Bus: articulated bus.
C-Cone: construction cone.
V-Trailer: vehicular trailer.
Some categories are excluded from the table due to the limited number of instances they contain. 
However, the average results consider all categories, even those that are omitted.
In this experiment, the input voxel size of our PoIFusion is (0.2m, 0.2m, 0.2m), the image resolution is $960\times 640$, and the image backbone is ResNet-50, following the setting of FSF~\cite{li2023fully}.
}
\vspace{-2mm}

\resizebox{\textwidth}{!}{%
\centering
\begin{tabular}{l|l|c|cccccccccccccccccccc}

\specialrule{1pt}{0pt}{1pt}
\toprule
    &
  \textbf{Methods} &
 \rotatebox{90}{Average} & 
 \rotatebox{90}{Vehicle} & 
 \rotatebox{90}{Bus} &
 \rotatebox{90}{Pedestrian} &
 \rotatebox{90}{Box Truck} &
 \rotatebox{90}{C-Barrel} &
 \rotatebox{90}{Motorcyclist} &
 \rotatebox{90}{MPC-Sign} &
 \rotatebox{90}{Motorcycle} &
 \rotatebox{90}{Bicycle} &
 \rotatebox{90}{A-Bus} &
 \rotatebox{90}{School Bus} &
 \rotatebox{90}{Truck Cab} &
 \rotatebox{90}{C-Cone} &
 \rotatebox{90}{V-Trailer} &
 \rotatebox{90}{Bollard} &
 \rotatebox{90}{Sign} &
 \rotatebox{90}{Large Vehicle} &
 \rotatebox{90}{Stop Sign} &
 \rotatebox{90}{Stroller} &
 \rotatebox{90}{Bicyclist} \\
\midrule
\multirow{7}{*}{mAP} & Far3D~\cite{jiang2024far3d} & 24.4 & - & - & - & - & - & - & - & - & - & - & - & - & - & - & - & - & - & - & - & - \\
& CenterPoint~\cite{centerpoint}        & 22.0 & 67.6 & 38.9 & 46.5 & 40.1 & 32.2 & 28.6 & 27.4 & 33.4 & 24.5 & 8.7 & 25.8 & 22.6 & 29.5 & 22.4 & 37.4 & 6.3 & 3.9 & 16.9 & 0.5 & 20.1 \\
& FSD~\cite{fsd}      & 28.2 & 68.1 & 40.9 & 59.0 & 38.5 & 42.6 & 39.7 & 26.2 & 49.0 & 38.6 & 20.4 & 30.5 & 14.8 & 41.2 & 26.9 & 41.8 & 11.9 & 5.9 & 29.0 & 13.8 & 33.4 \\
& VoxelNeXt~\cite{voxelnext}        & 30.5 & 72.0 & 39.7 & 63.2 & 39.7 & 64.5 & 46.0 & 34.8 & 44.9 & 40.7 & 21.0 & 27.0 & 18.4 & 44.5 & 22.2 & 53.7 & 15.6 & 7.3 & 40.1 & 11.1 & 34.9 \\
& FSF~\cite{li2023fully}    & 33.2 & 70.8 & 44.1 & 60.8 & 40.2 & 50.9 & 48.9 & 28.3 & 60.9 & 47.6 & 22.7 & 36.1 & 26.7 & 51.7 & 28.1 & 41.1 & 12.2 & 6.8 & 27.7 & \textbf{25.0} & 41.6 \\
& CMT~\cite{cmt} & 36.1 & 71.9 & 41.5 & 61.2 & 38.4 & 62.2 & 58.2 & 40.3 & 55.1 & 40.3 & 24.7 & 47.3 & 23.6 & 59.5 & 23.8 & 49.6 & \textbf{28.4} & 6.8 & \textbf{54.7} & 4.6 & 50.2 \\
& \cellcolor{mediumgray} PoIFusion (ours) 
& \cellcolor{mediumgray}\textbf{40.6} 
& \cellcolor{mediumgray}\textbf{77.6} 
& \cellcolor{mediumgray}\textbf{49.4} 
& \cellcolor{mediumgray}\textbf{70.6} 
& \cellcolor{mediumgray}\textbf{40.7} 
& \cellcolor{mediumgray}\textbf{75.0} 
& \cellcolor{mediumgray}\textbf{67.1} 
& \cellcolor{mediumgray}\textbf{44.0} 
& \cellcolor{mediumgray}\textbf{66.9} 
& \cellcolor{mediumgray}\textbf{55.1} 
& \cellcolor{mediumgray}\textbf{26.9} 
& \cellcolor{mediumgray}\textbf{49.2} 
& \cellcolor{mediumgray}\textbf{32.5} 
& \cellcolor{mediumgray}\textbf{64.6} 
& \cellcolor{mediumgray}\textbf{31.9} 
& \cellcolor{mediumgray}\textbf{59.0} 
& \cellcolor{mediumgray}28.1 
& \cellcolor{mediumgray}\textbf{8.0} 
& \cellcolor{mediumgray}44.0 
& \cellcolor{mediumgray}11.4 
& \cellcolor{mediumgray}\textbf{55.1} \\

\midrule 
\multirow{7}{*}{CDS} & Far3D~\cite{jiang2024far3d} & 18.1 & - & - & - & - & - & - & - & - & - & - & - & - & - & - & - & - & - & - & - & - \\
& CenterPoint~\cite{centerpoint}        & 17.6 & 57.2 & 32.0 & 35.7 & 31.0 & 25.6 & 22.2 & 19.1 & 28.2 & 19.6 & 6.8 & 22.5 & 17.4 & 22.4 & 17.2 & 28.9 & 4.8 & 3.0 & 13.2 & 0.4 & 16.7 \\
& FSD~\cite{fsd}          & 22.7 & 57.7 & 34.2 & 47.5 & 31.7 & 34.4 & 32.3 & 18.0 & 41.4 & 32.0 & 15.9 & 26.1 & 11.0 & 30.7 & 20.5 & 30.9 & 9.5 & 4.4 & 23.4 & 11.5 & 28.0 \\
& VoxelNeXt~\cite{voxelnext}                & 23.0 & 57.7 & 30.3 & 45.5 & 31.6 & 50.5 & 33.8 & 25.1 & 34.3 & 30.5 & 15.5 & 22.2 & 13.6 & 32.5 & 15.1 & 38.4 & 11.8 & 5.2 & 30.0 & 8.9 & 25.7 \\
& FSF~\cite{li2023fully}            & 25.5 & 59.6 & 35.6 & 48.5 & 32.1 & 40.1 & 35.9 & 19.1 & 48.9 & 37.2 & 17.2 & 29.5 & 19.6 & 37.3 & 21.0 & 29.9 & 9.2 & 4.9 & 21.8 & \textbf{18.5} & 32.0 \\
& CMT~\cite{cmt} & 27.8 & 62.2 & 33.6 & 46.8 & 30.8 & 47.3 & 47.6 & \textbf{30.2} & 43.1 & 29.8 & 18.9 & 38.4 & 16.9 & 42.5 & 17.1 & 34.5 & 21.1 & 5.0 & \textbf{43.0} & 3.2 & 40.4 \\
& \cellcolor{mediumgray} PoIFusion (ours) 
& \cellcolor{mediumgray}\textbf{31.6} 
& \cellcolor{mediumgray}\textbf{66.5} 
& \cellcolor{mediumgray}\textbf{40.8} 
& \cellcolor{mediumgray}\textbf{54.8} 
& \cellcolor{mediumgray}\textbf{33.0} 
& \cellcolor{mediumgray}\textbf{58.4} 
& \cellcolor{mediumgray}\textbf{54.6} 
& \cellcolor{mediumgray}28.7 
& \cellcolor{mediumgray}\textbf{55.0} 
& \cellcolor{mediumgray}\textbf{42.8} 
& \cellcolor{mediumgray}\textbf{20.4} 
& \cellcolor{mediumgray}\textbf{40.0} 
& \cellcolor{mediumgray}\textbf{24.7} 
& \cellcolor{mediumgray}\textbf{47.8} 
& \cellcolor{mediumgray}\textbf{23.5} 
& \cellcolor{mediumgray}\textbf{42.3} 
& \cellcolor{mediumgray}\textbf{21.6} 
& \cellcolor{mediumgray}\textbf{5.8} 
& \cellcolor{mediumgray}35.0 
& \cellcolor{mediumgray}8.7 
& \cellcolor{mediumgray}\textbf{42.8} \\

\bottomrule
\specialrule{1pt}{1pt}{0pt}
\end{tabular}
}

\label{tab:sota_argo}
\end{table*}

\subsection{Comparison with State-of-the-art Methods}\label{sec4.3:main_result}
\subsubsection{Results on nuScenes} 
In Table~\ref{tab:overall},
we make a comprehensive comparison on nuScenes 3D object detection benchmark. The compared methods are divided into three groups: camera-based methods (``C''), LiDAR-based methods(``L''), and multi-modal ones(``C+L''). All the compared multi-modal 3D object detection approaches are without temporal information. Both the test time augmentation and the model ensembling techniques are not used in this comparison. Broadly speaking, the multi-modal methods outperform the methods that leverage only one single modality, validating the benefits of integrating the semantic-intensive RGB image and location-aware LiDAR point clouds.

With the same image resolution and the same voxel size, our proposed PoIFusion achieves state-of-the-art performance on the nuScenes dataset. Specifically, equipped with ResNet-50, our method achieves 73.2\% NDS / 71.2\% mAP on the nuScenes validation set. Upgrading the image backbone to a more powerful Swin-T~\cite{swin}, the performance improves to 73.6\% NDS and 71.7\% mAP. We also submit the inference results to the official test server, and it reports that our PoIFusion with Swin-T image backbone achieves 74.9\% NDS and 73.4\% mAP, which makes an absolute improvement of 0.8\% NDS and 1.4\% mAP over the previous best results~\cite{cmt}.

Compared to unified-view approaches~\cite{bevfusion-damo, bevfusion-mit, uvtr}, our PoIFusion method preserves modal-specific information more effectively, leading to a significant improvement over the representative BEVFusion model~\cite{bevfusion-mit}, with gains of 2.0\% in NDS and 3.2\% in mAP on the test set. Additionally, in contrast to recent work such as ObjectFusion~\cite{cai2023objectfusion}, which generates object-centric features by fusing regional features through RoI Pooling, our method leverages multi-modal fusion at adaptive Points of Interest (PoIs), resulting in an improvement of 1.6\% in NDS and 2.4\% in mAP. The use of PoIs enhances the flexibility of sampling locations and enables fine-grained feature fusion, contributing to the superior performance of our approach.

\subsubsection{Results on Argoverse2} 
To study the performance of long-range 3D object detection, which is of significance to ensure safe driving, we further validate our approach on Argoverse2 dataset. The performance comparison is presented in Table~\ref{tab:sota_argo}. Compared to nuScenes, the long-tail category distribution of Argoverse2 emphasizes the importance of semantic-sensitive image features, while the precise localization of such a large detection range heavily relies on LiDAR sensor. Such facts make multi-modal fusion a promising choice for better 3D object detection. As shown in the table, our method consistently outperforms other competitors to deal with the challenge of long-range 3D object detection. Remarkably, the proposed PoIFusion sets a state-of-the-art record on this benchmark, achieving 40.6\% mAP and 31.6\% CDS. Note that the performance gap between our method and the strongest competitor CMT~\cite{cmt} on Argoverse2 is more significant than that of nuScenes, further demonstrating the advantage of our PoI-based fusion to extract object-relevant features for addressing more challenge detection tasks.

\begin{table}[t]
\footnotesize
\renewcommand{\arraystretch}{1.15}
\setlength{\tabcolsep}{5pt}
\caption{\textbf{Performance comparison of 3D multi-object tracking on nuScenes validation split} in terms of AMOTA (\%) and AMOTP (\%). }
 \resizebox{0.49\textwidth}{!}{%
 \centering
        \begin{tabular}{lcc}
        \hline
        \textbf{Methods} &  AMOTA $\uparrow$ & AMOTP $\downarrow$  \\
        \shline
    CenterPoint~\cite{centerpoint} &  63.7 & 60.6  \\
    VoxelNeXt~\cite{voxelnext} &  70.2 & 64.0 \\
    TransFusion~\cite{transfusion} & 71.8 & 60.3  \\
    BEVFusion (MIT)~\cite{bevfusion-mit}  & 72.8 & 59.4  \\
    ObjectFusion~\cite{cai2023objectfusion}  & 74.2 & 54.3  \\
    \rowcolor{mediumgray} PoIFusion(ours)& \textbf{75.1} & \textbf{50.7} \\
        \hline
        \end{tabular}
         }
	\label{tab:mot_result}
\end{table}

\subsubsection{Extended Results for Multiple Object Tracking}
In addition to the main results of 3D object detection, we evaluate the generalizability of our method on the 3D multiple object tracking task of the nuScenes dataset. Specifically, we follow CenterPoint~\cite{centerpoint} to adopt the ``tracking-by-detection'' scheme that offline links the detection boxes into tracking tubelets and to evaluate the performance in terms of AMOTA and AMOTP. We would like to clarify that the compared methods, except VoxelNeXt~\cite{voxelnext}, all follow the same ``tracking-by-detection'' scheme. As shown in Table~\ref{tab:mot_result}, our PoIFusion works the best (75.1\% AMOTA and 50.7\% AMOTP) among the compared methods. This experiment shows that our better detection results can facilitate the downstream task, \emph{i.e.}, 3D multiple object tracking, which is also an important component in the perception system of an autonomous-driving vehicle.

\begin{table}[t]
  \centering
  \footnotesize
  \caption{\textbf{Latency comparison.} The latency is evaluated on nuScene, with batch size set as 1. ``$\dagger$'': accelerated with Flash-Attention~\cite{dao2022flashattention} (NVIDIA V100 GPU is not compatible with the Flash Attention operator). Top-2 entries are with \textbf{bold font}.}
  
  \renewcommand{\arraystretch}{1.05}    
  \setlength{\tabcolsep}{6.2pt}
  \resizebox{0.49\textwidth}{!}{%
  \begin{tabular}{y{150}x{30}x{30}}
  \toprule
    \multirow{2}{*}{\textbf{Methods}}   & \multicolumn{2}{c}{Latency (ms)} 
\\
     & A100 & V100 \\
    \midrule
    TransFusion~\cite{transfusion}  & 153.8 & 190.3  \\
    BEVFusion (MIT)~\cite{bevfusion-mit}   & \textbf{113.9} & \textbf{135.5}  \\
    DeepInteraction~\cite{deepinteraction}  & 204.1 & 344.8\\
    CMT~\cite{cmt}   &  210.8 & 387.1 \\
    CMT$^\dagger$~\cite{cmt}  &  159.7 & - \\
    \rowcolor{mediumgray} PoIFusion (ours)   & \textbf{115.2} & \textbf{163.7} \\
    \bottomrule
    \end{tabular}
    }
    
    \label{tab:runtime_comparison}
    \vspace{-0.3cm}
\end{table}

\subsubsection{Latency Comparison}
Furthermore, we conduct the latency comparison among the recent works. For a fair comparison, we re-evaluate the latency on an NVIDIA A100 and an NVIDIA V100 GPU. Since the implementation of the voxelization operation will not influence the detection results but becomes a confounder for runtime evaluation, we exclude the costed time of voxelization in this comparison. As shown in Table~\ref{tab:runtime_comparison}, the latency of our PoIFusion is 115.2ms on an NVIDIA A100 GPU, which is comparable to the fastest method, \emph{i.e.}, BEVFusion (MIT), which takes 113.9ms to process each sample.
This comparison further demonstrates the potential of our proposed PoIFusion to serve as a strong baseline for future investigation and application.

\begin{table}[t]
\renewcommand{\arraystretch}{1.15}
\centering
\small
\caption{\textbf{Ablative experiments of components in the PoI multi-modal decoder}. The backbone networks are ResNet-50 and VoxelNet. Image resolution is set as $800\times448$, and the voxel size of point clouds is ($0.075m$,$0.075m$,$0.2m$). Our default setting is highlighted in \colorbox{mediumgray}{gray}.
}
\centering

\begin{minipage}{0.98\linewidth}{\begin{center}

\begin{tabular}{y{90}|x{60}x{60}}
\toprule
\textbf{Anchor points} & NDS (\%) & mAP (\%) \\
\midrule
Center only & 72.8  & 70.6  \\
\rowcolor{mediumgray} Center + corner & \textbf{73.2} &  \textbf{71.2}   \\
\bottomrule
\end{tabular}
\end{center}}
\subcaption{\textbf{The choice of anchor points.} Exploiting both the center and corner points as anchor points for points of interest generation yields better performance.}
\label{tab:ablation:anchor}
\end{minipage}
\\
\begin{minipage}{0.98\linewidth}{\begin{center}
\begin{tabular}{y{90}|x{60}x{60}}
\toprule
\textbf{PoI generation} & NDS (\%) & mAP (\%)\\
\midrule
Baseline & 72.4 & 70.1 \\
+ B.T. & 72.8  & 70.8 \\
\rowcolor{mediumgray} + B.T. \& P.S. & \textbf{73.2} &  \textbf{71.2}   \\
\bottomrule
\end{tabular}
\end{center}}
\subcaption{\textbf{Operations to derive PoIs.} The baseline in this experiment directly uses anchor points as PoIs. B.T.: box transformation. P.S.: point shift.} 
\label{tab:ablation:transformation}
\end{minipage}
\\
\centering
\begin{minipage}{0.98\linewidth}{\begin{center}
\begin{tabular}{y{90}|x{60}x{60}}
\toprule
\textbf{Fusion block}& NDS (\%) & mAP (\%) \\
\midrule
Static fusion & 71.8  &  69.3 \\
\rowcolor{mediumgray} Dynamic fusion & \textbf{73.2} &  \textbf{71.2}  \\
\bottomrule
\end{tabular}
\end{center}}
\subcaption{\textbf{Fusion block}. Dynamic fusion block produces adaptive parameters for the fusion layer, improving the performance.}
\label{tab:ablation:fusion_block}
\end{minipage}

\label{tab:ablations}
\vspace{-0.4cm}
\end{table}

\subsection{Experimental Analysis}\label{sec4.4:ablation}
In this section, we conduct several groups of controlled experiments to experimentally analyze the effects of each factor in our proposed PoIFusion framework. In these experiments, the image backbone network is ResNet-50 and the point cloud backbone network is VoxelNet. Unless specified, the image resolution is set as $800\times448$ and the voxel size is set as ($0.075m$,$0.075m$,$0.2m$). The models are all evaluated on the validation set of the nuScenes dataset.

\subsubsection{Components of PoI Multi-Modal Decoder} 
In this section, we provide a detailed analysis of the key components in the PoI multi-modal decoder. The experiments are structured to investigate three critical aspects: (a) the selection of anchor points, (b) the generation of PoIs from anchor points, and (c) the fusion strategy to integrate multi-modal information.

As shown in Table~\ref{tab:ablation:anchor}, the model that only uses the center point as the anchor to generate PoIs achieves 72.8\% NDS and 70.6\% mAP. By additionally involving the corner points as the anchor points, the NDS and mAP boosts to 73.2\% and 71.2\%, respectively. This comparison shows the benefits of representing the geometric property of query boxes.

In Table~\ref{tab:ablation:transformation}, we validate the operations to derive PoIs from anchor points. The baseline directly takes the anchor points as PoIs without adaptation, achieving 70.1\% mAP. By involving the holistic box-level transformation and further applying the point-wise shift, the performance boosts to 70.8 \% mAP and 71.2\% mAP, respectively. The online adjustment of PoIs ameliorates the misalignment in feature sampling and thereby boosts the detection result.

Table~\ref{tab:ablation:fusion_block} compares the effect of different fusion schemes, \emph{i.e.}, static fusion versus dynamic fusion. In static fusion, two conventional linear layers are exploited to process the concatenated multi-modal features at each PoI. The dynamic fusion scheme refers to our proposed on in Section~\ref{sec3:fusion}. Thanks to the adjustment of fusion layers' parameters for each object query, our dynamic fusion outperforms the static setting by 1.4\% NDS and 1.9\% mAP.

\begin{table}[t]

\caption{\textbf{Effect of leveraging different modalities}. Integrating RGB images and LiDAR point clouds remarkably boosts the detection result. }
\renewcommand{\arraystretch}{1.2}
\setlength{\tabcolsep}{4.8pt}
\centering
\small
\begin{tabular}{y{90}|x{60}x{60}}
\toprule
\textbf{Modality} & NDS (\%) & mAP (\%) \\
\midrule
Camera only & 40.6 & 29.4 \\
LiDAR only & 70.2 & 65.3 \\
\rowcolor{mediumgray} Camera + LiDAR & \textbf{73.2} &  \textbf{71.2} \\
\bottomrule
\end{tabular}
\label{tab:ablation:modality}
\end{table}

\subsubsection{Effect of Different Modalities} 
In this experiment, we train models with single-modality input following the same training setting as introduced in Section~\ref{sec4.1:exp_setup}.
This experiment is presented in Table~\ref{tab:ablation:modality}. The image-only baseline achieves 40.6\% NDS and 29.4\% mAP, and the LiDAR-only baseline achieves 70.2\% NDS and 65.3\% mAP. 
Integrating the multi-modal information remarkably boosts the performance to $73.2\%$ NDS and $71.2\%$ mAP.
Please note that the performance of our LiDAR-only baseline is comparable to that of TransFusion-L (70.1\% NDS and 65.1\% mAP), which is the LiDAR-only baseline of \cite{transfusion,bevfusion-damo,bevfusion-mit,deepinteraction,sparsefusion,cai2023objectfusion}, but our fusion model makes a more significant improvement. This comparison further validates the benefits of our proposed fusion at points of interest scheme to integrate complementary information from images and point clouds.

\subsubsection{Effect of the Input Resolution}
The input resolution (\emph{i.e.}, image resolution, and voxel size) is also one of the key factors that influence the detection result. In this experiment, we present the performance comparison under various input resolutions.

\noindent\textbf{Image Resolution.} As shown in Table~\ref{tab:ablation:img_resolution}, we enlarge the image resolution from $800\times448$ to $1600\times640$, which is the same as the input resolution of the strongest competitor CMT. Compared to our default setting, the detection performance is observed with an improvement of 0.5\% NDS and 0.7\% mAP. The larger resolution provides more detailed textual information from the RGB images and thus benefits multi-modal 3D object detection. It is worth noting that although increase the image resolution can definitely improve the performance, we have not increase it in our default setting, which ensures fair comparison with others.

\begin{table}[t]
\renewcommand{\arraystretch}{1.2}
\centering
\small
\caption{\textbf{Ablative experiments of the input resolution}. The backbone networks are ResNet-50 and VoxelNet. Our default setting is highlighted in \colorbox{mediumgray}{gray}.
}
\centering

\begin{minipage}{0.98\linewidth}{\begin{center}

\begin{tabular}{y{90}|x{60}x{60}}
\toprule
 \textbf{Image resolution} & NDS (\%) & mAP (\%) \\
\midrule
\rowcolor{mediumgray} $800\times448$ & 73.2  & 71.2  \\
$1600\times640$ & \textbf{73.7} &  \textbf{71.9}   \\
\bottomrule
\end{tabular}
\end{center}}
\subcaption{\textbf{Image Resolution}. Increasing the image resolution strengthens the textual information provided by images.}
\label{tab:ablation:img_resolution}

\end{minipage}
\\
\begin{minipage}{0.98\linewidth}{\begin{center}
\begin{tabular}{y{90}|x{60}x{60}}
\toprule
\textbf{Voxel size} & NDS (\%) & mAP (\%) \\
\midrule
(0.125, 0.125, 0.2) & 71.6  & 69.1 \\
(0.1, 0.1, 0.2) & 72.3 & 69.8 \\
\rowcolor{mediumgray} (0.075, 0.075, 0.2) & \textbf{73.2} &  \textbf{71.2}  \\
\bottomrule
\end{tabular}
\end{center}}
\subcaption{\textbf{Voxel Size}. A smaller voxel size corresponds to a larger resolution of the point cloud feature, facilitating the localization information for 3D object detection.}
\vspace{-0.3cm}
\label{tab:ablation:voxel_size}
\end{minipage}
\\

\label{tab:ablations2}
\end{table}

\noindent\textbf{Voxel Size.} In Table~\ref{tab:ablation:voxel_size}, we further analyze the influence of the voxel size by setting the voxel size as (0.125m, 0.125m, 0.2m), (0.1m, 0.1m, 0.2m) and (0.075m, 0.075m, 0.2m). As shown in the table, the model with a smaller voxel size works better than that with a larger one. Smaller voxel size results in a larger resolution of the point cloud feature map, which facilitates more precise localization information.

\begin{figure*}[t]
  \centering
  \includegraphics[width=0.98\linewidth]{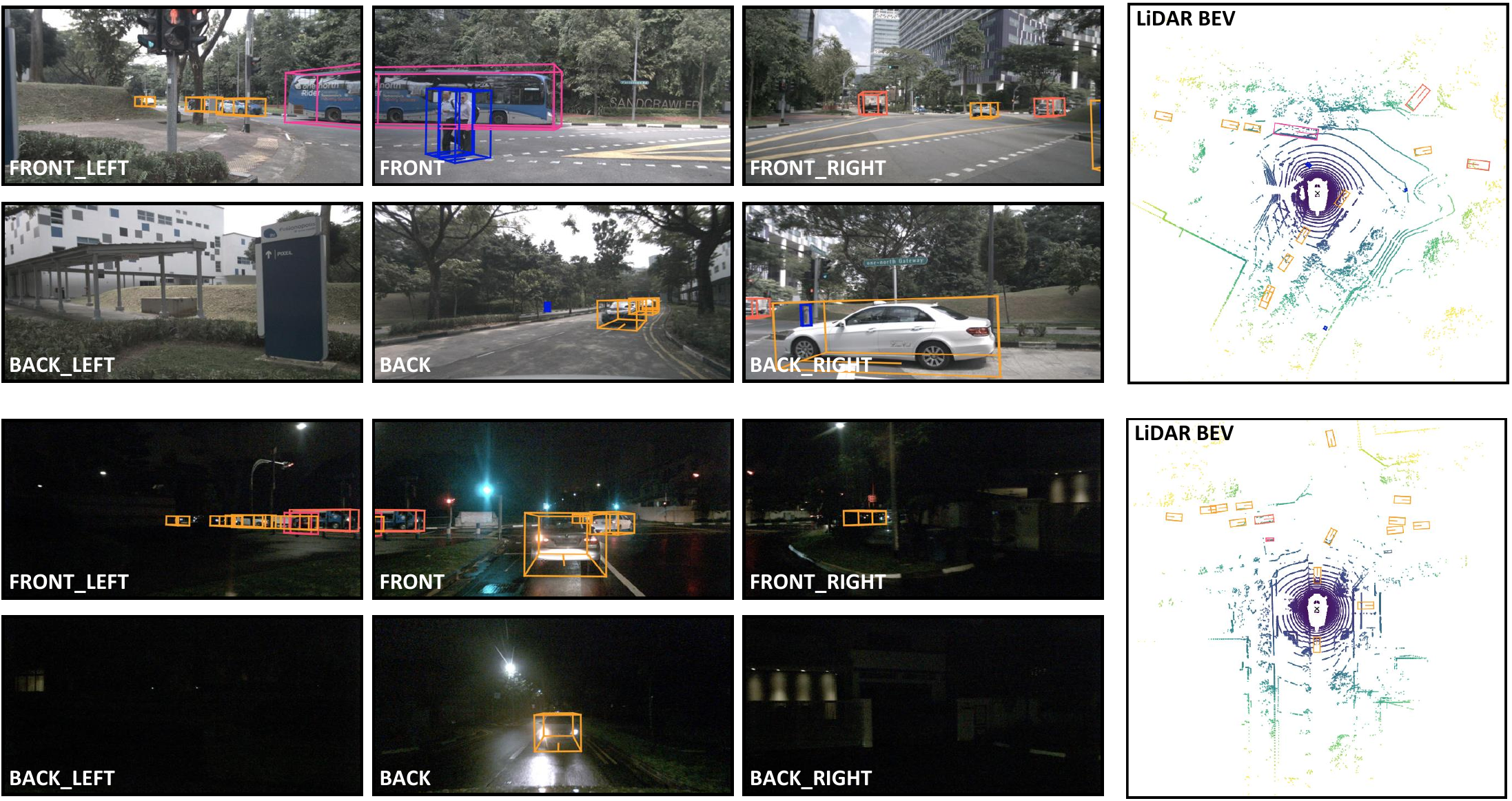}
  \caption{\textbf{Qualitative results on the nuScenes validation set.} One example from a sunny day and another example from a rainy night are presented. Our PoIFusion precisely detects 3D objects under varying weather and lighting. We use different colors for different categories.}
  \label{fig:qualitative_example}
\end{figure*}

\begin{table}[t]
\renewcommand{\arraystretch}{1.2}
	 \caption{\textbf{Performance comparison between using shared parameters and unshared parameters for the PoI multi-modal decoder.} Our default setting is highlighted in \colorbox{mediumgray}{gray}.}
	\centering
        \small
        \begin{tabular}{y{90}|x{60}x{60}}
        \toprule
        \textbf{Parameters} & NDS (\%) & mAP (\%) \\
        \midrule
        Unshared parameters & \textbf{73.3} & 70.9\\
        \rowcolor{mediumgray} Shared parameters & 73.2 & \textbf{71.2}\\
        \bottomrule
        \end{tabular}
    
	\label{tab:shared_parameters}
\end{table}

\subsubsection{Shared Parameters VS. Unshared Parameters for the PoI Multi-Modal Decoder} 
Our PoI multi-modal decoder is applied iteratively, with the option to use either shared or unshared parameters. The performance comparison between these two configurations is shown in Table~\ref{tab:shared_parameters}, where both models demonstrate comparable results. However, employing shared parameters reduces the overall model size, making it more efficient. Consequently, we adopt the shared-parameter configuration as the default setting for the PoI multi-modal decoder.

\begin{table}[t]
	 \caption{\textbf{Different weathers and lighting conditions.} The reported performance is compared in terms of mAP (\%) on the nuScenes validation set. We report the result of our PoIFusion with the Swin-T image backbone network, which is consistent with MVP~\cite{mvp} and BEVFusion~\cite{bevfusion-mit}.}
\renewcommand{\arraystretch}{1.2}
	\centering
	\resizebox{0.49\textwidth}{!}{
	\begin{tabular}{l|c|cccc}
  \hline
  {mAP (\%)}
         & {Modality} & {Sunny} & {Rainy} & {Day} & {Night} \\
        \hline
        {BEVDet}  & C & 32.9 & 33.7 & 33.7 & 13.5 \\
        
        {Centerpoint} & L & 62.9 & 59.2 & 62.8 & 35.4 \\
        \hline
        {MVP} & C+L & 65.9 & 66.3 & 66.3 & 38.4 \\
        {BEVFusion} & C+L & 68.2 & 69.9 & 68.5 & 42.8 \\
        \rowcolor{mediumgray} PoIFusion (ours) & C+L & \textbf{71.1} & \textbf{70.2} & \textbf{71.3} & \textbf{45.0} \\
        \hline

	\end{tabular}
	}

	\label{tab:abl_daynight}
\end{table}

\subsection{Robustness Analysis}\label{sec4.5:robust_analysis}

Besides the accuracy and efficiency, the capability of working under different environments and 
corruptions is also what we expected for a desirable perception module in autonomous driving vehicles. In this section, we present a series of experiments to examine the robustness of our PoIFusion framework.

\subsubsection{Different Weathers and Lighting Conditions}
The 3D object detection system on autonomous driving vehicles is supposed to be capable of working under varying lighting and weather conditions.  
In this experiment, we follow~\cite{bevfusion-mit} to divide the validation set of the nuScenes dataset into two pairs of subsets: sunny/rainy and day/night for comprehensive quantitative evaluation. The evaluated performance in terms of mAP (\%) is summarized in Table~\ref{tab:abl_daynight}. 
In general, the multi-modal 3D object detectors are not likely to be affected by the rainy weather. The changing of lighting condition significantly affects the detection performance, \emph{i.e.}, the mAP on the night subset is much lower than that of the day subset. 
Among the compared methods, our PoIFusion achieves the best performance under different weather and lighting conditions, validating the robustness of fusion at points of interest.

Moreover, we visualize the qualitative results of detecting 3D objects under different weather and lighting conditions in Figure~\ref{fig:qualitative_example}. We present one example from a sunny day and another example from a rainy night. These examples further give an intuitive understanding that PoIFusion can robustly perform 3D object detection under varying weather and lighting conditions

\begin{table}[t]
\small
\renewcommand{\arraystretch}{1.15}
\setlength{\tabcolsep}{10.8pt}
\caption{\textbf{Sensor misalignment.} Performance comparison on the impact of translation offsets due to calibration errors. The reported performance is compared in terms of mAP (\%) on the nuScenes validation set. We report the result of our PoIFusion with the ResNet-50 image backbone network.}
 \centering
        \begin{tabular}{l|cc}
        \hline
        Misalignment offset (m)  & NDS (\%) &mAP (\%) \\
        \shline
        0.0 & 73.2 & 71.2  \\
        0.2 & 73.0 & 71.1  \\
        0.4 & 72.9 & 70.8  \\
        0.6 & 72.7 & 70.5  \\
        0.8 & 72.3 & 70.0  \\
        1.0 & 72.0 & 69.4 \\
        \hline
        \end{tabular}
	\label{tab:translation}
\end{table}

\subsubsection{ Robustness Against Sensor Misalignment}
In real applications, sensors can be misaligned due to physical impacts, installation errors, time-related drift, or some other unexpected reasons.
This experiment investigates the robustness of PoIFusion against sensor misalignment. Specifically, we follow~\cite{transfusion} to randomly add translation offsets to the calibration matrix at inference. The errors are formulated as uniformly distributed noise, with the maximum offset value varying from $0.2m$ to $1.0m$. As shown in Table~\ref{tab:translation}, even when the maximum translation offset is $1.0m$, our method can achieve 72.0\% NDS and 69.4\% mAP, still improving our LiDAR-only baseline with 1.8\% NDS and 4.1\% mAP. This experiment demonstrates the robustness of our fusion paradigm against sensor misalignment.

\subsubsection{Robustness Against Sensor Failure} In addition to sensor misalignment, we further explore a more terrible corruption, \emph{i.e.}, sensor failure. 

\vspace{+0.1cm}
\noindent \textbf{Camera Failure.} To validate our method under camera failure cases, we randomly drop several images by filling the corresponding images as all zeros. The results with different numbers of dropped images are presented in Table~\ref{tab:camera_failure}. The performance keeps dropping along with the increasing number of cameras that fail to work. Even with half of the cameras blocked, our PoIFusion achieves 68.6\% mAP, obviously better than the LiDAR-only method. Moreover, in the extreme case that all of the cameras are not working, our method still works a little bit better than the LiDAR-only baseline, showing the benefit of multi-modal joint training.

\begin{table}[t]
\small
\renewcommand{\arraystretch}{1.15}
\caption{\textbf{Camera failure.} Performance comparison on the impact of dropped images due to camera failure.}
\setlength{\tabcolsep}{13.8pt}
 \centering
        \begin{tabular}{l|cc}
        \hline
        \# Dropped Images  & NDS (\%) &mAP (\%) \\
        \shline
        0 & 73.2 & 71.2  \\
        1 & 72.8 & 70.3  \\
        3 & 72.0 & 68.6  \\
        6 & 70.2 & 65.6  \\
        \hline
        \end{tabular}
	\label{tab:camera_failure}
\end{table}

\begin{table}[t]
\small
\renewcommand{\arraystretch}{1.15}
\caption{\textbf{LiDAR failure.} Performance comparison on the impact of discarded point clouds within a sector due to LiDAR failure. }
 \centering
 \setlength{\tabcolsep}{13.8pt}
        \begin{tabular}{l|cc}
        \hline
        Discarded Sectors  & NDS (\%) &mAP (\%) \\
        \shline
        $0^\circ$ & 73.2 & 71.2  \\
        $6^\circ$ & 72.7 & 70.2  \\
        $12^\circ$ & 72.0 & 68.9  \\
        $18^\circ$ & 71.3 & 67.6  \\
        $24^\circ$ & 70.7 & 66.4  \\
        \hline
        \end{tabular}
	\label{tab:lidar_failure}
\end{table}

\vspace{+0.1cm}
\noindent \textbf{LiDAR Failure.} Furthermore, to simulate the failure of a LiDAR sensor, we randomly discard point cloud data within a specified sectorial region based on the azimuthal angle. 
As shown in Table~\ref{tab:lidar_failure}, in general, the impact of LiDAR failure is more pronounced than that of camera failure, since the information captured by the LiDAR sensor is essential for precise localization. 
Notwithstanding the total absence of point cloud data within a $24^\circ$ sector, the mAP achieved by our method is still 1.1\% higher than the LiDAR-only baseline. These two experiments demonstrate the robustness of our proposed PoIFusion against Sensor Failure.

\section{Conclusion}
In this work, we introduce PoIFusion, a query-based multi-modal 3D object detector that leverages multi-modal feature sampling and fusion at strategically generated points of interest. PoIFusion preserves modal-specific information by maintaining the original views of both image and point cloud features, while the use of points of interest allows for fine-grained fusion and a flexible representation of object features. Our approach establishes a new state-of-the-art on the challenging nuScenes dataset, all while maintaining efficient runtime performance. Furthermore, experimental results highlight the robustness of PoIFusion to sensor misalignment and failure. We believe that PoIFusion will serve as a robust baseline with significant potential for future research and deployment in the field.
\section{Data Availability}
The nuScenes dataset~\cite{nus} and Argoverse2 dataset~\cite{argo2} used in this study are well-recognized benchmarks which are public available. Specifically, nuScenes dataset can be accessed through  \url{https://www.nuscenes.org/} and Argoverse2 dataset can be accessed through \url{https://www.argoverse.org/av2.html}. We have not used extra private data in our experiments.

\bibliographystyle{spbasic}

\bibliography{sn-bibliography}

\end{document}